\documentclass[sigconf]{acmart}
\usepackage{algorithm}
\usepackage{algorithmic}
\usepackage{array}
\newcommand{\T}{\mathcal{T}}

\newcommand{\D}{\mathcal{D}}
\newcommand{\cc}{\mathbf{c}}
\newcommand{\x}{\mathbf{x}}
\newcommand{\z}{\mathbf{z}}

\newcommand{\h}{\mathbf{h}}
\newcommand{\Real}{\mathbb{R}}
\usepackage{multirow}
\usepackage[toc,page]{appendix}
\newcommand{\tabincell}[2]{\begin{tabular}{@{}#1@{}}#2\end{tabular}}  
\newcommand\blfootnote[1]{%
  \begingroup
  \renewcommand\thefootnote{}\footnote{#1}%
  \addtocounter{footnote}{-1}%
  \endgroup
}
\AtBeginDocument{%
  \providecommand\BibTeX{{%
    \normalfont B\kern-0.5em{\scshape i\kern-0.25em b}\kern-0.8em\TeX}}}

\setcopyright{none}
\renewcommand\footnotetextcopyrightpermission[1]{} 
\begin{document}
\fancyhead{}

\title{HetMAML: Task-Heterogeneous Model-Agnostic Meta-Learning for Few-Shot Learning Across Modalities}

\author{Jiayi Chen, Aidong Zhang}
\affiliation{%
  \institution{University of Virginia}
\country{}
}
\email{{jc4td, aidong}@virginia.edu}

\renewcommand{\shortauthors}{Trovato and Tobin, et al.}

\begin{abstract}
  Most of existing gradient-based meta-learning approaches to few-shot learning assume that all tasks have the same input feature space. However, in the real world scenarios, there are many cases that the input structures of tasks can be different, that is, different tasks  may  vary  in  the number  of  input  modalities or data types.  Existing meta-learners cannot handle the heterogeneous task distribution (HTD) as there is not only global meta-knowledge shared across tasks but also type-specific knowledge that distinguishes each type of tasks. To deal with task heterogeneity and promote fast within-task adaptions for each type of tasks, in this paper, we propose HetMAML, a task-heterogeneous model-agnostic meta-learning framework, which can capture both the type-specific and globally shared knowledge and can achieve the balance between knowledge customization and generalization. Specifically, we design a multi-channel backbone module that encodes the input of each type of tasks into the same length sequence of modality-specific embeddings. Then, we propose a task-aware iterative feature aggregation network which can automatically take into account the context of task-specific input structures and adaptively project the heterogeneous input spaces to the same lower-dimensional embedding space of concepts. Our experiments on six task-heterogeneous datasets demonstrate that HetMAML successfully leverages type-specific and globally shared meta-parameters for heterogeneous tasks and achieves fast within-task adaptions for each type of tasks.\blfootnote{Preprint. Accepted by CIKM21.}
\end{abstract}




\maketitle

\section{Introduction}

\begin{figure*}[t]
\centering
\includegraphics[width=2.05\columnwidth]{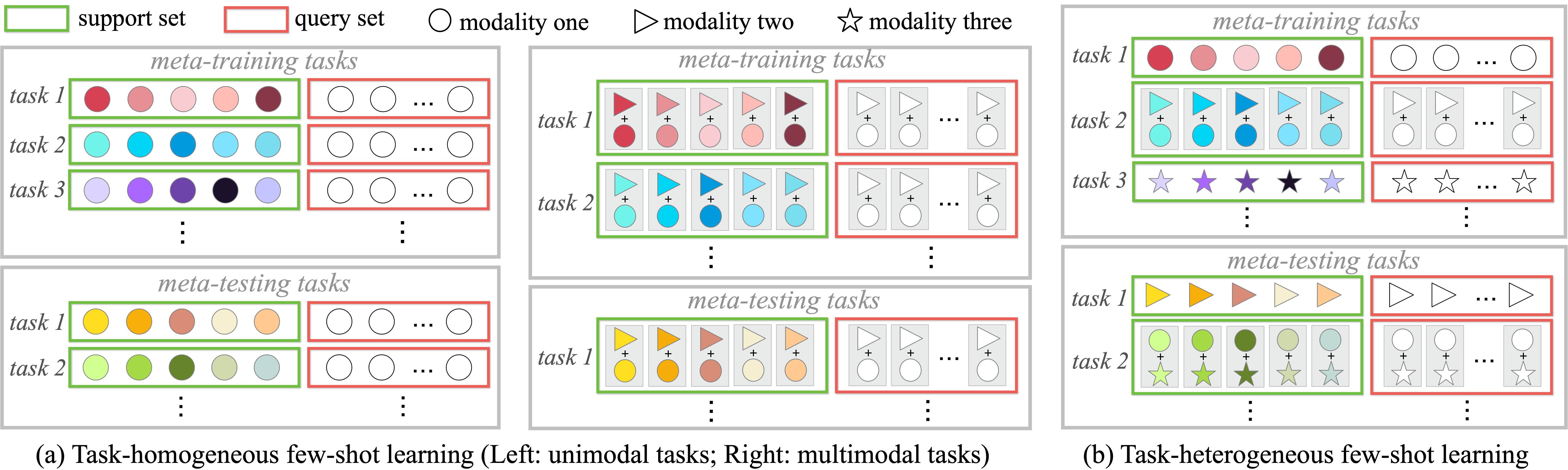}
\vspace{-0.1in}
\caption{Comparison of task-homogeneous few-shot learning and task-heterogeneous few-shot learning. 
Compared with task-homogeneous few-shot learning, heterogeneous tasks vary in terms of the input spaces, data types and the structure of input data, and thus will perform different within-task adaptions. }
\label{HTD}
\end{figure*}

Humans can quickly learn new concepts from few examples by incorporating prior knowledge and context.
Just like humans, \textit{few-shot learning} (FSL) is the task of learning new concepts with a small number of labelled samples. \textit{Meta-learning} is one prominent family of approaches to address FSL problems, which focuses on learning how to learn: how to generalize meta-knowledge from a distribution of tasks and utilize it to rapidly adapt to new tasks from the same task distribution. 
Recently, many \textit{gradient-based meta-learning} methods have been proposed to handle FSL tasks~\cite{finn2017model,yoon2018bayesian,vuorio2019multimodal,DBLP:conf/iclr/RusuRSVPOH19,DBLP:conf/iclr/MishraR0A18,zintgraf2019fast}.
Most of these methods are model-agnostic meta-learners built upon MAML \cite{finn2017model}, aiming to find a single set of model parameters from similar tasks so that a small number of gradient updates will lead to good performance on future new tasks.

\begin{figure*}[t]
\centering
\includegraphics[width=2.05\columnwidth]{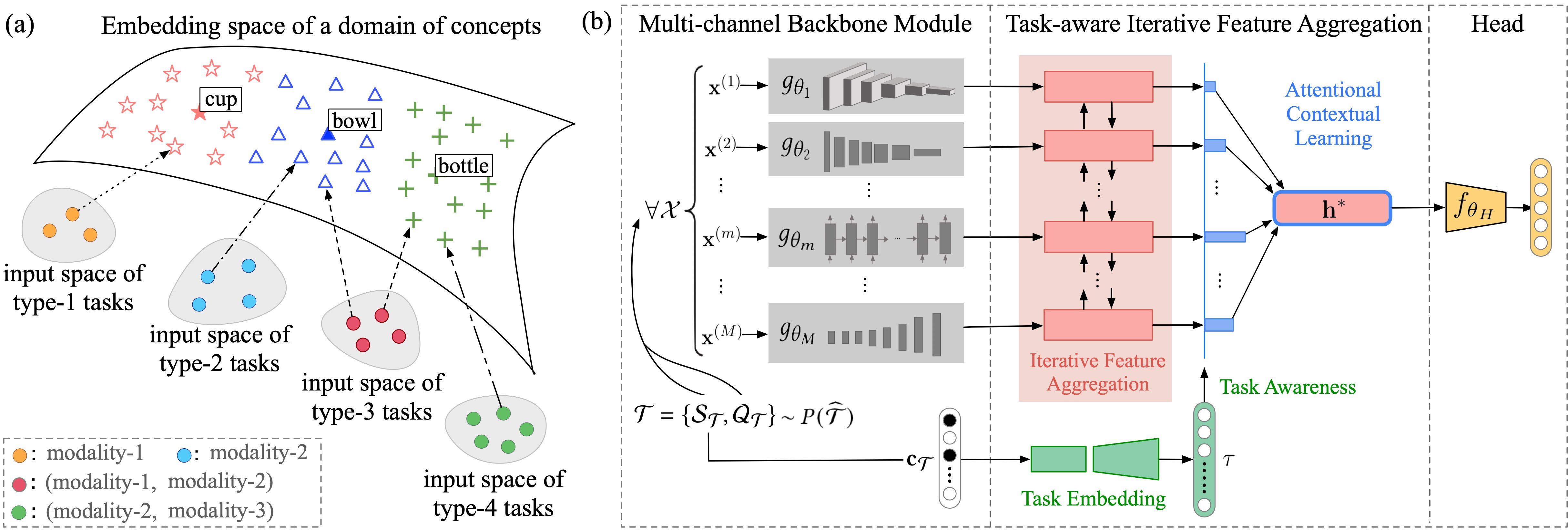}
\caption{(a) An example of heterogeneous tasks sharing the concept domain but having different input spaces. There are a total of four types of tasks; each grey area indicates the input space of a certain type of tasks and the feature space shared by all data samples in these tasks. 
Colored filled circles represents unimodal or multimodal data samples; different colors within grey areas indicate that the samples have different structures, which are listed in the bottom-left box. 
Note that here we omit the id of samples to clearly show their structure differences. 
The arrow lines represent task-aware projections, meaning that data samples are mapped from different feature spaces to a unified concept space.
(b) The proposed HetMAML framework.
}
\label{fig1}
\end{figure*}

However, most existing gradient-based meta-learning methods assume the task distribution is \textit{homogeneous}, that is, all tasks from either the same concept domain (e.g., a single dataset \cite{finn2017model}) or a combination of different domains (e.g., multiple datasets \cite{vuorio2019multimodal}) have the same input feature space--for example, the inputs of all tasks are same-dimensional images.  
Such an assumption limits the generalization ability of these methods to handle more complex and \textit{heterogeneous task distribution} (HTD): different tasks may vary in the structure of input data or the number of input modalities.
For example, in few-shot emotion classification, some tasks may focus on recognizing people's emotion from the video modality, while other tasks may predict on language scripts, recorded voices, or any combination of the three modalities. 
Figure \ref{HTD}(a) shows examples of task-homogeneous FSL, which assume identical data structures in all tasks.
Figure \ref{HTD}(b) shows an example of task-heterogeneous  FSL. Compared with the homogeneous FSL in Figure \ref{HTD}(a), heterogeneous tasks vary in terms of data type and the structure of input. Figure \ref{fig1}(a) illustrates another example of task-heterogeneous FSL containing four types of tasks, where we show each \textit{type} of tasks is associated with a specific \textit{input space} (see the grey areas in the figure). The input space of a type of tasks refers to the feature space shared by all the training/validation samples within these tasks. 


In this paper, we propose a gradient-based meta-learning approach to FSL under heterogeneous task distributions. The novel task-heterogeneous few-shot learning problem is inevitable in many real-world low-data scenarios. Considering tasks can be either unimodal or multimodal with respect to their input data sources, there are mainly two cases of FSL across modalities.
First, different modalities may convey the same semantics of a concept, and it is feasible to \textit{jointly learn multiple types of unimodal tasks} if they share knowledge in the same concept domain. For instance, the image of an animal and the caption of it (text modality) both can distinguish this animal from others. Simultaneously learning image classification and text classification tasks that share the same concept domain could help us to build up the knowledge of animal concepts.
Second, 
integrating multiple views of a concept or using auxiliary modalities
usually achieves better performance than learning with a single modality \cite{xing2019adaptive,pahde2020multimodal}. In this multimodal case, a problem neglected by existing literature is that due to the potential data missingness or data corruptness in some few-data scenarios, many tasks may not have a complete set where all modalities are always available \cite{lahat2015multimodal}. Hence, it is uncertain that all these multimodal tasks have the same input structure, and we have to \textit{jointly learn multiple types of multimodal tasks, where different tasks may have different combinations of modalities}.
For example, in Figure \ref{fig1}(a) where there are four types of tasks, while type-2 tasks only have the modality-2 data available, type-2 and type-3 tasks have the auxiliary modality-1 and modality-3, respectively. 
For both cases, it is beneficial to learn the heterogeneous tasks across modalities due to the following reasons: 1) jointly learning multiple views of tasks could help to build up the knowledge of the concept domain; 2) we can generalize reliable semantic feature extractors due to the implicit cross-modal alignment; and 3) the mechanism of training a unified meta-learner over heterogeneous tasks is more efficient than training separate models for each type of tasks as there are global meta-parameters that can be shared across different types of tasks.

With the heterogeneous task distribution (HTD), tasks share the same concept domain but have different input spaces (as shown in Figure \ref{fig1}(a)).  As for meta-learning, while heterogeneous tasks could globally share some common knowledge (e.g., knowledge of the concept domain and early-stage feature embedding), there is also
\textit{type-specific knowledge}, locally shared by each type of tasks, which can reflect the different concept learning mechanisms of different types of tasks.
Therefore, the key challenge of dealing with task heterogeneity is how to automatically 
reckon with
type-specific information to promote fast within-task adaptions for each type of tasks. That is, we pursue \textit{how to customize the globally shared meta-learner} for each type of tasks.

We address two issues in this challenge. 
First, the meta-learner for HTD should be able to 
balance \textit{generalization} (i.e., the meta-knowledge globally shared across tasks) and \textit{customization} (i.e., the type-specific meta-knowledge able to customize the global knowledge) based on each task's specific input space. The difficulty is that, given a new task sampled from HTD whose input structure is \textit{unclear}, the meta-learner should automatically leverage type-specific and global knowledge to achieve the \textit{task-aware} projection: the mechanism that maps samples from the input space to the concept space should be different for different types of tasks, as illustrated in Figure \ref{fig1}(a) (arrow lines). Although one may convert the problem setting from heterogeneity into homogeneity using preprocessing strategies like imputation \cite{cai2018deep,shang2017vigan,tran2017missing}, this may introduce extra noise to the original task which has negative impacts on the performance, especially in low-data scenarios. Directly learning with different input spaces is necessary.
Second, due to the inconsistency of input structures, tasks may use different model architectures and thus cannot share a single set of model parameters.
Although one can train separate homogeneous meta-learners \cite{finn2017model,vuorio2019multimodal} for each type of tasks, this strategy is not efficient as well as not effective: there are parameters shared by several types of tasks but trained repeatedly; the separate training will reduce the number of training tasks in each type which limits the ability of knowledge generalization; and, the knowledge shared between tasks is not well-explored. 

To tackle the above issues, we propose Task-\textbf{Het}erogeneous \textbf{M}odel-\textbf{A}gnostic \textbf{M}eta-\textbf{L}earning (HetMAML). The key idea of HetMAML is to leverage globally shared knowledge and type-specific knowledge to promote fast within-task adaptions for each type of tasks. We propose a three-module task-aware learning framework that can effectively capture both the global and type-specific meta-parameters and achieve the knowledge customization while simultaneously preserving knowledge generalization. 
Our main contributions are summarized as follows:

\begin{itemize}
\item We study a novel task-heterogeneous few-shot learning problem involving multiple input spaces, data types, and modality combinations. We show the importance of studying this real-world problem and define our setting of task heterogeneity. As far as we know, this is the first attempt to this problem. 

\item We introduce HetMAML, a model-agnostic meta-learning approach to task-heterogeneous few-shot learning. We propose a task-aware iterative feature aggregation network to promote the effectiveness of knowledge customization as well as adaption efficiency.

\item Our experimental results on six datasets demonstrate that HetMAML can effectively capture global and type-specific meta-parameters across heterogeneous tasks and fast adapt to different types of new tasks.

\end{itemize}







\section{Methodology}

\subsection{Problem Formulation}

In this section, we will provide an explicit formulation of the few-shot learning problem under heterogeneous task distribution. 

\textbf{Definition 1 (\textit{Heterogeneous Task Distribution}).}  Suppose we have a total of $M$ data sources (modalities) and the data samples in each task are either unimodal or multimodal. A \textit{heterogeneous task distribution} (HTD) $P(\widehat{\T})$ is defined as a mixture of $M'$ homogeneous task distributions $P(\T^{1}), P(\T^{2}),...,P(\T^{M'})$, where each $P(\T^{r})$ is associated with a specific input space, i.e., a certain combination of the $M$ modalities. Then there will be a total of $M'$ types of task instances sampled from the HTD, where $ 2\leq M'\leq 2^M-1$.


\textbf{Task-heterogeneous Meta-learning}. We are given a dataset $\D$ consisting of task instances sampled from a heterogeneous task distribution $P(\widehat{\T})$. The dataset $\D$ is divided into the meta-training $\D_{meta}^{trn}$ and the meta-testing $\D_{meta}^{tst}$ sets. Each meta-training task contains $N$ classes sampled from a set of classes $\mathcal{C}^{trn}$, while the classes of each meta-testing task are sampled from a disjoint set of new classes $\mathcal{C}^{tst}$. The goal of task-heterogeneous meta-learning is to find a set of meta-parameters from $\D_{meta}^{trn}$ which can rapidly adapt to future novel tasks $\D_{meta}^{tst}$.

Each task instance $\T_i \sim P(\widehat{\T})$ is composed of a within-task training set $\mathcal{S}_{\T_i}$ (\textit{support set}) and testing set $\mathcal{Q}_{\T_i}$ (\textit{query set}), each of which consists of a few \textit{unimodal} \textit{or} \textit{multimodal} data samples. The support set of a $N$-way $K$-shot classification task contains $K$ samples for each of $N$ classes
$\mathcal{S}_{\T_i} = \{(\mathcal{X}_{ink}, y_{ink}) | n=1...N, k = 1...K\}$,
where $\mathcal{X}_{ink}=(\x_{ink}^{(1)}, \x_{ink}^{(2)} ,..., \x_{ink}^{(M)})$  denotes the composite input of each sample and $y_{ink}$ is the label. 
The query set contains $T$ samples from the same $N$ classes $\mathcal{Q}_{\T_i} = \{(\mathcal{X}^*_{it}, y^*_{it}) | t=1...T\}$. 
Note that although  we use the notation $\mathcal{X}$ to include a full set of $M$ modalities for each sample, it still contains the information about which modalities are not available in this task. For example,
if the $m$th source does not exist in task $\T_i$, the source website or sensor will return a message indicating its unavailability, or the $m$th modality of each sample $\mathcal{X}$ in a task is not informative. 
Then, given a task $\T_i$, we are able to  recognize the input space of this task by computing a configuration vector $\cc_{\T_i}=[c_{\T_i}^{(1)} \cdots c_{\T_i}^{(M)}]^{\top}$ from the raw data in the support set $\mathcal{S}_{\T_i}$ (described in Section \ref{MBM}).







\subsection{Task-Heterogeneous Model-Agnostic Meta-Learning Framework }

 
A traditional task neural network can be viewed as a two-module architecture, consisting of a \textit{backbone} module (the earlier layers for feature extraction) and a \textit{head} module (the later layers for decision making such as classification).  Most existing model-agnostic meta-learners for homogeneous tasks are built upon such architecture.

Due to the existence of task heterogeneity, we design a three-module framework, HetMAML, as illustrated in Figure \ref{fig1}(b), consisting of three modules: a multi-channel backbone, a single-channel head module, and an additional intermediate module--the task-aware feature aggregation network (TFAN), which can \textit{adaptively} transform the composite information produced by the backbone into a single hidden representation fed to the head. 



\subsubsection{Multi-Channel Backbone Module}\label{MBM}

A challenge in this problem is that the input feature space of samples is type-specific; each type of tasks has a specific combination of modalities and hence the feature extraction schemes between different tasks can be different. 
For example, while image classification tasks use convolutional neural networks (CNNs) to extract features, other tasks that involve sequential data may use recurrent neural networks (RNNs). 
Despite the feature embedding being type-specific, each modality's feature extractor can be shared by several types of tasks. 
In order to enable simultaneous learning over heterogeneous tasks, our backbone module should be able to effectively model both the shared and type-specific feature extractors. To achieve this goal, we propose a \textit{multi-channel backbone module}, where each channel is a modality-specific feature extractor. 
Basically, if two tasks both have a modality in their input spaces, they can share the same feature extractor's meta-parameters for embedding this modality.

Given a task $\T$, it is easy to identify the input space of this task based on the data samples in the support set $\mathcal{S}_{\T}$. We can obtain a task-specific vector $\cc_{\T}=[c_{\T}^{(1)}, c_{\T}^{(2)},…, c_{\T}^{(M)}]^{\top}$ to represent the input space, where $c_{\T}^{(m)}=1$ denotes that the $m$th modality is available in this task ($c_{\T}^{(m)}=0$ denotes missing). 
Considering the missing modalities are not informative, $c_{\T}^{(m)}$ is set to 0 if and only if the $m$th modalities of all the samples in $\mathcal{S}_{\T}$ are similar such that
 $\text{Var}^{(m)}=\sum_{n,k} ||{\x_{nk}}^{(m)} - \overline{\x}^{(m)}||^2/NK < \epsilon$,
where $\overline{\x}^{(m)}=\sum_{n,k} {\x_{nk}}^{(m)}/NK$ and $\epsilon$ is a threshold. 

Then, suppose $\Theta_B=\{\theta_1, \theta_2,...,\theta_M\}$ is the parameter set of the multi-channel backbone module, and $g_{\theta_m}(\cdot;\theta_m)$ is the feature extracting network for modality-$m$ with parameter $\theta_m$. Based on the pre-computed $\cc_{\T}$ which indicates the input space of the task,  for any sample
$\mathcal{X}=(\x^{(1)}, \x^{(2)}, ...., \x^{(M)})$ in the task $\T$, each modality can be embedded as $\z^{(m)} = g_{\theta_m}(\x^{(m)}; \theta_m)$ if $c_{\T}^{(m)}=1$; or, $\z^{(m)}=\mathbf{0} $ if $c_{\T}^{(m)}=0$.
Note that there will be no gradient update for those channels with $c_{\T}^{(m)}=0$.

The multi-channel backbone module will produce $M$ embedded vectors: $\z^{(1)}, \z^{(2)}, ..., \z^{(M)}$. Each embedding vector $\z^{(m)} \in \Real^{F_1}$ represents a piece of \textit{modality-specific information}. Note that despite the architecture differences between channels, all modalities are encoded onto the same $F_1$-dimensional semantic space, reducing the impact of input statistical divergence of different types of tasks. 



\subsubsection{Task-Aware Feature Aggregation for Heterogeneous Tasks}



The set of encoded modality-specific information $\{\z^{(m)}\}_{m=1}^M$ produced by the multi-channel backbone is not ready for decision making. One may consider simply concatenating all modality-specific vectors and directly feeding it to the head neural networks. However, this strategy is not effective due to the following two reasons. First, a concatenated vector is large and sparse--its dimension is $M\cdot F_1$ and many modalities that are absent in the task are represented by zero. This will result in a large feature mapping network at the beginning of the head module, which is not feasible in few-shot learning setting. Second, when there are multimodal types of tasks, the multiple input modalities can be highly-interacted \cite{zadeh2017tensor}, hence it is necessary to consider their potential interactions before decision making. These reasons motivate us to consider to effectively combine and condense all the modality-specific features. 
Thus, we insert an intermediate \textit{feature aggregation} module $\Theta_A$ between the backbone module and head module, which can transform the set of modality-specific information produced by the backbone into a single lower-dimensional and non-sparse vector that represents the \textit{task-specific information}. 

In large-scale multimodal learning \cite{zadeh2017tensor,liu2017multi}, aggregating multiple features is a general topic. However, there are two challenges when multiple features are aggregated in a few-data scenario with an irregular input structure configuration.
In other words, the multi-feature aggregation in a task-heterogeneous few-shot setting should satisfy two prerequisites: 1) \textbf{Efficiency of feature aggregation}. According to \cite{lahat2015multimodal}, multiple modalities contain complementary information, and effective aggregation should be able to combine them while leveraging their potential interactions. Although techniques such as Tensor Fusion Network (TFN) \cite{zadeh2017tensor} have achieved effective interaction modeling, they basically require large-scale training data due to extensive parameters. We found that, in few-shot learning, such a large model tends to lead to over-fitting and requires more training episodes during adaption. Also, simple methods such as the concatenation are faster and smaller in model size but less powerful in exploring multimodal interactions. Correspondingly, in a heterogeneous few-data scenario, we should balance the \textit{adaption efficiency} and the \textit{exploration of multimodal interactions}. 
2) \textbf{Task awareness of feature aggregation}. Homogeneous few-shot learning focuses on learning the transferable knowledge generally shared across tasks (i.e., \textit{generalization}). In contrast, in our heterogeneous few-shot setting, each task has a specific combination of modalities so that how to perform multimodal interactions can be very different between tasks. For example, while image classification tasks only analyze image data, a bimodal task containing both image and text should consider image-text interactions.  
The feature aggregation network shared by all types of tasks should be aware of each task's specific input structure so that it can adaptively perform multimodal interactions. In other words, in addition to generalization, we also should learn how to customize the globally shared meta-learner (i.e., \textit{customization}). That is, under a heterogeneous task distribution, the feature aggregation network should balance  \textit{generalization} and \textit{customization}. 

Motivated by the two prerequisites, we propose Task-aware Feature Aggregation Network (TFAN), which contains two components: 1) Iterative Feature Aggregation based on the bidirectional recurrent neural network and 2) Attentional Contextual Learning, to achieve the task-awareness of feature aggregation. 

\textbf{\textit{Iterative Feature Aggregation Network}}. As mentioned above, to achieve an effective feature aggregation in low-data scenarios, we pursue a balance between the exploration of multimodal interactions and the adaption speed. 
We found that the existing large-scale multimodal fusion technique \cite{zadeh2017tensor,zadeh2018memory} as well as the straightforward feature combinations such as concatenation cannot effectively achieve this goal. 
Alternatively, we propose to employ the framework of bidirectional recurrent neural network (BRNN) \cite{schuster1997bidirectional} to iteratively combine the information of multiple modalities channel by channel while training parameters for processing only one channel at each step. 

Suppose the set of multi-channel encoded modality-specific information $\{\z^{(m)}\}_{m=1}^M$ constructs a sequence with $M$ components $Z=(\z^{(1)}, \z^{(2)} \cdots \z^{(M)})$,
where we consider each piece of modality-specific information $\z^{(m)}$ as a \textit{token} as in text embedding. 
Using the bidirectional long-short term memory (BiLSTM) \cite{hochreiter1997long} network as an example of BRNN, in text embedding, BiLSTM learns a representation for each token by combining with the other words as context. Similarly, given the sequence $Z$, each modality-specific embedding $\z^{(m)}$ can be combined with the others through the 
bidirectional modality-by-modality \textit{iterative} calculation process:
\begin{equation}
\begin{array}{lcl}
\overrightarrow{\h}^{(m)} = \overrightarrow{\text{LSTM}}(\z^{(m)},\overrightarrow{\h}^{(m-1)})\\
\overleftarrow{\h}^{(m)} = \overleftarrow{\text{LSTM}}(\z^{(m)},\overleftarrow{\h}^{(m+1)})\\
\h^{(m)} = \overrightarrow{\h}^{(m)}\oplus \overleftarrow{\h}^{(m)} \in \Real^{F_2},
\end{array}
\end{equation}
where $\oplus$ denotes concatenation and $F_2$ is the dimension of hidden state. The iterative feature aggregation network will produce $M$ hidden states: $\mathcal{H}=\{\h^{(m)}\}_{m=1}^M$. 
Each hidden state $\h^{(m)}$ is considered as a view of \textit{aggregated multimodal information} combining all modality-specific information in $Z$. 
%

The idea of iterative aggregation is to encode multimodal features modality by modality while exploring their interactions step by step. According to the forward pass function of an LSTM's unit
\begin{equation}\label{lstm}
\begin{array}{lcl}
v_f^{m} =\text{sigmoid}(W_{f} \z^{(m)} + U_{f} \textbf{h}^{(m-1)} + b_f) \\
v_i^{m} = \text{sigmoid}(W_{i} \z^{(m)} + U_{i} \textbf{h}^{(m-1)} + b_i) \\
v_o^{m} = \text{sigmoid}(W_{o} \z^{(m)} + U_{o} \textbf{h}^{(m-1)} + b_o) \\
\widetilde{v_c}^{m} = \text{tanh}(W_{c} \z^{(m)} + U_{c} \textbf{h}^{(m-1)} + b_c) \\
v_c^{m} = v_f^{m} \circ v_c^{m-1} + v_i^{m} \circ \widetilde{v_c}^{m} \\
\textbf{h}^{(m)} = v_o^{m} \circ \text{tanh}(v_c^{m}),
\end{array}
\end{equation}
where $\{W_{f}, U_{f}, b_f\}$, $\{W_{i}, U_{i}, b_i\}$, $\{W_{o}, U_{o}, b_o\}$, and $\{W_{c}, U_{c}, b_c\}$ are the LSTM parameters of the forget gate, the input gate, the output gate, and the memory cell, respectively, at each iterative step, the memory cell's update mechanism enables the feature interaction between each new coming feature $\z^{(m)}$ and previous aggregated information $\textbf{h}^{(m-1)}$. Note that Eq. (\ref{lstm}) is for each direction.

As for adaption efficiency, the number of trainable parameters of the iterative feature aggregation is $O((F_1+F_2)F_2)$, where $F_1$ and $F_2$ are dimensions of the embedding vector of each modality and the hidden state of the recurrent network, respectively.  Compared with the highly-interacted feature aggregation methods designed for large-scale data, such as the multimodal outer-product containing $O((F_1)^MF_2)$ parameters \cite{zadeh2017tensor}, our model reduces the computation complexity of fusion and thus requires less training episodes and adapts faster. Also, compared with early fusion methods such as \cite{perez2013utterance,morency2011towards,liu2017multi} containing $O(MF_1F_2)$ parameters, iterative aggregation is more powerful in exploring multimodal interactions while the computation complexity is unrelated with $M$. In addition, as we also face the task-heterogeneity challenge, iterative aggregation works better in the task-heterogeneous few-shot learning environment.



\paragraph{\textbf{Attentional Contextual Learning for Task-aware Feature Aggregation}}  Under a heterogeneous task distribution, we pursue the balance between generalization (globally shared knowledge) and customization (task-specific knowledge). In other words, the feature aggregation network should be aware of each task's specific input structure, and learn how to explore task-specific knowledge to effectively customize the globally shared feature integration process.
We propose the attentional contextual learning to achieve such balance, the task-awareness of feature aggregation.

First, since our definition of task heterogeneity is the configuration or availability of a set of input modalities,
the sequence $Z=(\z^{(1)}, \z^{(2)} \cdots \z^{(M)})$ can reflect the specific knowledge for the given task's input structure--each modality has its own position in this sequence and the absent modalities can be considered as a special token. In other words, we can view the input structure of tasks as the ``context" of the sequence $Z$, which can distinguish different types of tasks. In text embedding, BRNN models the sequential data by leveraging context dependencies in the input sequence (\textit{context-aware}) \cite{zhou2016attention}. Similarly, when the BRNN-based iterative feature aggregation encodes the sequence of modalities, it can take into account the input structure (context) of the given task, which reflects task-specific knowledge.

Besides, to further customize the iterative feature aggregation network using task-specific knowledge, we propose an external meta-network $g_{ \phi}(\cdot;\phi)$, which learns a task embedding vector $\tau$ for each given task to represent the task-specific knowledge as
\begin{equation}
\mathbf{\tau} = g_{ \phi}(\cc_{\T}; \phi) \in \Real^{F_3},
\end{equation}
where $F_3$ is the dimension of task embedding. We let the task embedding $\mathbf{\tau}$ represent the type-specific knowledge that reflects each task's specific input space and distinguishes different tasks; hence,  $\mathbf{\tau}$ is embedded from the $\cc_{\T}$ calculated in Section \ref{MBM}. 
Then, we utilize $\mathbf{\tau}$ as extra knowledge to customize the globally shared meta-parameters in the feature aggregation network.
Specifically, we employ a modified attention mechanism \cite{DBLP:journals/corr/BahdanauCB14} to facilitate the customization. The final task-specific information is represented by 
\begin{equation}
\h^{*}=\sum_{m=1}^M A_m \h^{(m)} \in \Real^{F_2},
\end{equation}
where $\{A_m\}_{m=1}^M$ are attention coefficients indicating the role of each modality with respect to the type-specific knowledge $\tau$: 
%
\begin{equation}
    A_m = \frac{\text{exp}(\mathbf{v}^T \text{tanh}(\mathbf{W}_h [ \h^{(m)} \oplus \mathbf{\tau}]))}{\sum_l \text{exp}(\mathbf{v}^T \text{tanh}(\mathbf{W}_h [ \h^{(l)} \oplus \mathbf{\tau}]))},
\end{equation}where $ \oplus$ denotes concatenation operation,  and $\mathbf{v}\in\Real^{F_3}$ and $ \mathbf{W}_h\in\Real^{F_3\times (F_2 + F_3)}$ are learnable parameters of the attention module. The attention mechanism incorporates the type-specific knowledge $\mathbf{\tau}$ into the feature aggregation, so that it can \textit{adaptively} perform the within-task adaption for different types of tasks. 

Overall, our task-aware feature aggregation network (TFAN) transforms the multi-channel modality-specific information into task-specific information. Network parameters in this module include the iterative feature aggregation $\theta_{iter}$, the task-embedding $\phi$, and the attention module $\{\mathbf{v}, \mathbf{W}_h\}$. 


\subsubsection{Single-Channel Head Module} 


In the previous stage, TFAN adaptively projects samples from different input spaces onto a unified concept space--the generated task-specific representations of each sample in different type of tasks are in the same concept space (see arrow lines in Figure \ref{fig1}(a)). As for final decision making, we use a single-channel head module $f_{\theta_H}(\cdot;\theta_H )$. For few-shot classification, the architecture of the head module is a multi-layer classifier followed by a softmax layer, i.e., $\text{softmax}(f_{\theta_H}(\h^*; \theta_H))$. 

\subsection{Training Procedure}

HetMAML is built upon the model-agnostic meta-learning (MAML) framework \cite{finn2017model, raghu2019rapid}, which solves a\textit{ bilevel optimization} problem to find an \textit{initialization} $\Theta_0$ for a neural network as the meta-parameters. 
This training procedure assumes that the single meta-initialization $\Theta_0$ is an appropriate generalization of prior knowledge for all tasks. It requires tasks to be homogeneous in terms of the size of labelled data, the classes number, the input structure, and so on.

Unlike homogeneous MAMLs, our task-heterogeneous model leverages extra task-specific knowledge to perform slightly different adaption between different types of tasks, that is, finding a balance between generalization and customization. Therefore, we divide meta-parameters into two disjoint sets: \textit{internal parameters} and \textit{external parameters}.

\subsubsection{Internal and External Parameters}
We define \textit{internal parameters} (IP) $\Theta$ as the meta-parameters, which take the gradient updates during the inner-loop optimization, to perform the adaption to each specific task. The internal parameters of HetMAML include the meta-initialization of the iterative feature aggregation network and the head module, i.e., $\Theta=\{\theta_{iter}, \theta_H\}$. The task-aware iterative feature aggregation projects input features onto the concept space, by considering the input context of each task. Hence $\theta_{iter}$ may vary among different tasks and need to adapt to each task. The head module (classifier) $\theta_H$ learns decision boundaries within the concept space. Since different tasks have different concepts, the parameters for dividing the concept space is specific for each task.

\textit{External parameters} (EP) $\Phi$ are defined as the meta-parameters, which pass through the neural network but does not need to adapt to each task during inner-loop optimization. The external parameters of HetMAML include the parameters of multi-channel backbone module, the task embedding network as well as the attention mechanism used for customization, i.e., $\Phi=\{\theta_B, \phi, \mathbf{v}, \mathbf{W}_h\}$. The backbone $\theta_B$ can be shared across tasks as it performs early-stage feature embedding of each data source. Meta-parameters in the task embedding network $\phi$ aim to capture the knowledge of how to distinguish different types of tasks, based on the structure property of each given task. $\phi, \mathbf{v}, $ and $\mathbf{W}_h$ embeds each task's input structure and utilize it to customize the internal parameters during adaption. 

\subsubsection{Bilevel Optimization}
The overall training procedure of HetMAML is described in Algorithm \ref{alg:algorithm}. Following \cite{finn2017model, raghu2019rapid}, the \textit{outer loop} updates the meta-initialization of internal parameters $\Theta_0$ and the external parameters $\Phi$ to enable fast adaptation over a batch of task instances. 
The inner loop takes the outer-loop  $\Theta_0$ and $\Phi$, and, separately for each task, performs a few gradient updates of internal parameters over the labelled examples in the support set, while freezing external parameters. 

Formally, let $\Theta_i'$ signify $\Theta$ for task $\T_i$ during the inner-loop optimization, and let the initial $\Theta_i'=\Theta_0$. In the inner-loop adaption, during each gradient update, we compute
\begin{equation}
\label{inner}
\Theta_{i}' \xleftarrow{} \Theta_{i}' - \alpha \nabla_{\Theta_{i}'}\mathcal{L}_{\T_i}(f(\mathcal{X}; \Theta_{i}', \Phi), y;\mathcal{S}_{\T_i}),
\end{equation}
where $f(\cdot)$ is the forward function of HetMAML network, and $\mathcal{L}_{\T_i}(\cdot;\mathcal{S}_{\T_i})$ is the loss on the support set of task $\T_i$. 

Separately for each task, after a fixed number of inner-loop updates, we obtain the adapted parameter $\Theta_i'(\Theta_0)$,  which is dependent on meta-initialization $\Theta_0$. Then, the outer-loop optimization updates $\Theta_0$ and $\Phi$ over a batch of task instances:
\begin{equation}\label{outer1}
    \Theta_0 \xleftarrow{} \Theta_0 - \beta \nabla_{\Theta_0} \sum_{\T_i \sim p(\widehat{\T})} \mathcal{L}_{\T_i}(f(\mathcal{X}^*; \Theta_i'(\Theta_0), \Phi), y^*;\mathcal{Q}_{\T_i})
\end{equation}
\begin{equation}\label{outer2}
\Phi \xleftarrow{} \Phi - \beta \nabla_{\Phi} \sum_{\T_i \sim p(\widehat{\T})} \mathcal{L}_{\T_i}(f(\mathcal{X}^*; \Theta_i'(\Theta_0), \Phi),y^*;\mathcal{Q}_{\T_i}), 
\end{equation}
where $\mathcal{L}_{\T_i}(\cdot;\mathcal{Q}_{\T_i})$ is the loss on the query set of task $\T_i$.








\begin{algorithm}[t]
   \caption{Training Procedure of HetMAML}
  \begin{algorithmic}[1]
    \STATE \textbf{Requires:} Heterogeneous task distribution $P(\widehat{\T})$
    \STATE \textbf{Requires:} Learning rates $\alpha$, $\beta$
    \STATE Randomly initialize internal parameters $\Theta$.
    \STATE Randomly initialize external parameters $\Phi$.
    \STATE //outer-loop optimization
    \WHILE{not done}
        \STATE Sample batches of heterogeneous tasks $\T_i \sim P(\widehat{\T})$
        \STATE // inner-loop optimization
        \FORALL{$i$}
            \STATE Obtain data $\{\mathcal{S}_{\T_i}, \mathcal{Q}_{\T_i}\}$ for each task $\T_{i}$.
            \STATE Obtain task structure  $\cc_{\T_i}$ from each $\mathcal{S}_{\T_i}$ as Section \ref{MBM}.
            \STATE Obtain task embedding $\tau_i=g(\cc_{\T_i}; \phi)$ using external $\phi$.
            \STATE Compute adapted internal parameters with a fixed number of steps w.r.t. the $NK$ examples from $\mathcal{S}_{\T_i}$ as in Eq.(\ref{inner}). 
            \STATE Evaluate 
            $\mathcal{L}_{i}(f(\mathcal{X^*}; \Theta'_{i}, \Phi), y^*;\mathcal{Q}_{\T_i})$ w.r.t. $T$ samples of $\mathcal{Q}_{\T_i}$.
        \ENDFOR
        \STATE Update initialization of internal parameters $\Theta_0$ as Eq.(\ref{outer1}).
        \STATE Update external parameters $\Phi$ as Eq.(\ref{outer2}).
    \ENDWHILE
    \STATE \textbf{return:} $\Theta_0$  and  $\phi$
  \end{algorithmic}
  \label{alg:algorithm}
\end{algorithm}

\section{Experiments}

In this section, we compare HetMAML with state-of-the-art baselines on six task-heterogeneous few-shot datasets.

\subsection{Datasets}

Since we define a new task-heterogeneous few-shot learning problem, we constructed our datasets from existing multimodal datasets. 



\subsubsection{Selection of the Source Multimodal Datasets} 

One prerequisite for the source multimodal dataset is that the total number of classes $|\mathcal{C}|$ should be large. First, we have to split $\mathcal{C}$ into two disjoint sets of class set, $\mathcal{C}^{trn}$ and $\mathcal{C}^{tst}$, for constructing $\mathcal{D}_{meta}^{trn}$ and $\mathcal{D}_{meta}^{tst}$, respectively. Then, the class set $\mathcal{C}_i$ of each task instance is sampled from $\mathcal{C}$. To construct a meta dataset containing a proper number of task instances (for training meta-learner),  where the $N$ classes in different task instances do not overlap too much, a large number of total classes is necessary. Another requirement is that, in the source multimodal dataset, the number of subjects (multimodal samples) for each class should also be large to make sure each data sample will not repeat frequently over tasks. Since some multimodal datasets with three or more modalities such as emotion classification datasets have a small number of classes and subjects, we did not consider using them as the source.

We chose the following two datasets as our \textit{source} multimodal datasets: 
1) Caltech-UCSD-Birds 200-2011 (\textsc{CUB-200}) \cite{WahCUB_200_2011} dataset contains 11,788 images for 200 bird species, including black footed albatross, bobolink, fish crow, and so on. Each image is annotated with a vocabulary of 312 attributes (e.g., crown color, belly color, eye color, etc.), which  we use as the text modality.  We used the two data sources from the CUB-200 dataset:  the \textit{image modality} $\x^{image} \in \Real^{3\times256\times256}$, and the \textit{text modality} $\x^{text}\in \Real^{312}$.  
2) \text{ModelNet40} \cite{wu20153d} is a 3D object detection dataset containing 12,311 3D CAD shapes covering 40 common categories, including airplane, bathtub, bookshelf, bottle, bowl, cone, cup, and so on. Following \cite{feng2019hypergraph}, we use it as a bimodal dataset where the two modalities are the two views of shape representation extracted from two pre-trained networks: Multi-view Convolutional Neural Network (MVCNN) \cite{su2015multi} and Group-View Convolutional Neural Network (GVCNN) \cite{feng2018gvcnn}. Specifically, the two data sources from the ModelNet40 are: \textit{modality-X1} (MVCNN representation) $\x^{mvcnn}\in \Real^{4096}$ and \textit{modality-X2} (GVCNN representation) $\x^{gvcnn}\in \Real^{2048}$. For simplicity, we will use ``X1'' and ``X2'' to represent the two modalities of ModelNet40.

\begin{table}[t]
\centering
    \begin{tabular}{p{1.8cm} c c p{2.7cm}}
  \toprule    & \scriptsize$|C^{trn}|/|C^{tst}|$\footnotesize & $M'$  & Task Input Structures \\ 
  \hline

    \multirow{2}{*}{\tabincell{c}{\textit{het}ModelNet-1}} &  \multirow{2}{*}{30/10} & \multirow{2}{*}{2} &     {Type 1:  X1 } \\
     & &  &{Type 2:  X2 } \\
    \hline
    \multirow{2}{*}{\tabincell{c}{\textit{het}ModelNet-2}} &   \multirow{2}{*}{30/10} & \multirow{2}{*}{2}  &     {Type 1:  X1 } \\
      &  & & {Type 2:  (X1,X2) } \\
    \hline
    \multirow{2}{*}{\tabincell{c}{\textit{het}ModelNet-3}} &  \multirow{2}{*}{30/10}  & \multirow{2}{*}{2}  &     {Type 1:  X2 } \\
      &  & & {Type 2:  (X1,X2) } \\
    \hline
    \multirow{3}{*}{\tabincell{c}{\textit{het}ModelNet-4}} &  \multirow{3}{*}{30/10}  & \multirow{3}{*}{3}  &     {Type 1:  X1 } \\
      &  & &{Type 2:  X2 } \\
      &  & &{Type 3:  (X1,X2) } \\
      \hline
          \multirow{2}{*}{\tabincell{c}{\textit{het}CUB200-1}} &    \multirow{2}{*}{150/50} &  \multirow{2}{*}{2}  &     {Type 1:  image } \\
      &  &  & {Type 2:  text } \\
      \hline
    \multirow{2}{*}{\tabincell{c}{\textit{het}CUB200-2}} &    \multirow{2}{*}{150/50} &  \multirow{2}{*}{2}  &     {Type 1:  image } \\
      &  &  & {Type 2:  (image, text) } \\
      
\bottomrule
\end{tabular}
\caption{Statistics of the six task-heterogeneous few-shot datasets (X1: MVCNN modality, X2: GVCNN modality).}
\label{miniModelNet}
\end{table}

\subsubsection{Construction of Task-Heterogeneous Datasets}

From source datasets, we constructed six task-heterogeneous few-shot datasets in total, which were built up as follows. First, we split the class set into two subsets, i.e., $\mathcal{C}=\{\mathcal{C}^{trn}, \mathcal{C}^{tst}\}$,  such that $|\mathcal{C}^{trn}|=\text{int}(3/4|\mathcal{C}|)$ and $\mathcal{C}^{tst}=\mathcal{C} \setminus \mathcal{C}^{trn}$. 
Then, we generated task instances for each of the meta-train and meta-test datasets: $|\mathcal{D}_{meta}^{trn}|=N_{trn}$ tasks for the meta-train dataset and $|\mathcal{D}_{meta}^{tst}|=N_{tst}$ tasks for the meta-test dataset.
Specifically, for constructing each task $\T$ in $\mathcal{D}_{meta}^{trn}$, we randomly selected $N$ classes from $C_{trn}$ and then selected $K$ samples from each class to form the support set; then, we selected \textit{other} $K_q$ labelled samples for each class to form the query set.
Finally, we split each of $\mathcal{D}_{meta}^{trn}$ and $\mathcal{D}_{meta}^{tst}$ into $M'$ groups; each group contains one type of tasks--we deleted certain modalities in tasks following the configuration in Table \ref{miniModelNet}. 
Statistics of the six constructed task-heterogeneous datasets are described below.

\textbf{\textit{het}CUB200-1 and 2}. Using the multimodal samples in CUB200, we constructed $N_{trn}=4000$ and $N_{tst}=1000$ task instances with $K_q=12$. Each data sample in CUB200 is a pair of (image, text) modalities. We considered the text modality as either the auxiliary modality for image or a view of sample that provides useful information under missing-image conditions. Therefore, we built two datasets; both has two types of tasks--the first type is unimodal tasks that only process images. In \textit{het}CUB200-1, tasks have either image or text modalities. In \textit{het}CUB200-2, the task heterogeneity was created by deleting the text modality from a half of tasks. 

\textbf{\textit{het}ModelNet-1 (2, 3, and 4)}. Using the multimodal samples in ModelNet40, we constructed $N_{trn}=4000$ and $N_{tst}=1000$ task instances with $K_q=12$. We considered different conditions of modality combinations--the modality-X1, the modality-X2, and a pair of modalities (X1,X2). Hence we could construct \textit{four} datasets from ModelNet40, whose input structures are listed in Table \ref{HTD}. 

\begin{table*}
\begin{center}
\begin{tabular}{lcccccccccccc}
\toprule
\multirow{3}{*}{Method} &\multicolumn{3}{c}{\textbf{\textit{het}ModelNet-1}}&\multicolumn{3}{c}{\textbf{\textit{het}ModelNet-2}}&\multicolumn{3}{c}{\textbf{\textit{het}ModelNet-3}}&\multicolumn{3}{c}{\textbf{\textit{het}ModelNet-4}}\\
\cmidrule(lr){2-4}
\cmidrule(lr){5-7}
\cmidrule(lr){8-10}
\cmidrule(lr){11-13}
& 5-way &5-way & 10-way  & 5-way & 5-way &10-way & 5-way & 5-way &10-way   & 5-way & 5-way &10-way       \\
& 1-shot & 5-shot & 1-shot & 1-shot & 5-shot& 1-shot   & 1-shot & 5-shot& 1-shot  & 1-shot & 5-shot& 1-shot   \\
\midrule

MAML  &  80.6 & 91.2  & 60.5 & 80.4 & 90.6  & 46.5 & 89.6  & 98.7  & 70.1 & 84.2 & 94.7  & 67.7\\
Multi-MAML(TF)  &  75.7 & 94.2  & 50.5 & 68.2  & 88.6   & 32.3 & 77.4  & 93.3 & 54.3 & 74.3 & 91.4  & 47.1 \\
Multi-MAML(BF) & 76.1  &  94.3 & 50.7 & 80.8  &  94.3   & 62.5  & 92.7  & 98.9 & 81.3  & 83.9 & 94.5 & 66.5  \\
\midrule
\textbf{HetMAML} (ours)   &  \textbf{84.5} & \textbf{94.7}  &  \textbf{72.8} & \textbf{85.7} & \textbf{95.2} & \textbf{71.9 } &\textbf{95.0} & \textbf{99.3}  & \textbf{90.1} & \textbf{92.3} & \textbf{95.3}  & \textbf{78.3} \\
\bottomrule
\end{tabular}
\end{center}
\caption{Few-shot classification accuracy (\%) on the meta-test splits of \textit{het}ModelNet-1 (2, 3, and 4) datasets.}
\vspace{-0.05in}
\label{results_modelnet}
\end{table*}

\begin{table}
\begin{center}
\begin{tabular}{lcccc}
\toprule
\multirow{3}{*}{Method} &\multicolumn{2}{c}{\textbf{\textit{het}CUB200-1}}&\multicolumn{2}{c}{\textbf{\textit{het}CUB200-2}}\\
\cmidrule(lr){2-3}
\cmidrule(lr){4-5}
& 5-way &5-way & 5-way  & 5-way   \\
& 1-shot & 5-shot & 1-shot & 5-shot \\
\midrule

MAML  &   43.9 & 60.2 & 52.7 & 69.5\\
Multi-MAML(TF)  & 41.3 & 63.1 & 39.1  & 51.2 \\
Multi-MAML(BF)  & 41.4 & 62.8 & 50.3  & 68.4 \\
\midrule
\textbf{HetMAML} (ours)    &\textbf{ 48.7} & \textbf{64.3} & \textbf{54.2 } & \textbf{70.7 } \\
\bottomrule
\end{tabular}
\end{center}
\caption{Few-shot classification accuracy (\%) on the meta-test splits of  \textit{het}CUB200-1 and \textit{het}CUB200-2 dataset.}
\vspace{-0.2in}
\label{results_cub}
\end{table}

\subsection{Baselines}

First, we compared HetMAML with  \textbf{MAML} \cite{finn2017model} designed for unimodal task distribution, which is limited to homogeneous tasks with the same input structure, and cannot be directly applied to our task-heterogeneous setting. We pre-processed the original heterogeneous inputs such that the processed inputs have the same feature space. The input dimension is fixed to that of the modality which has the largest input feature space.
Besides, similar to \cite{vuorio2019multimodal}, we used Multi-MAML baselines which consist of multiple MAML models trained separately on each type of tasks. We tested two versions of {Multi-MAML}s: \textbf{Multi-MAML (TF)} and \textbf{Multi-MAML (BF)}. 
For fair comparisons, we let each MAML also have an intermediate module: while Multi-MAML (TF) uses a Tensor Fusion Network \cite{zadeh2017tensor} for feature aggregation, Multi-MAML (BF) uses a BRNN network (i.e., BiLSTM) as in HetMAML.

\subsection{Results and Discussions}

All the experiments in this paper were conducted on a single-core GPU using Pytorch 3. 
In all experiments, we let 
$F_1=128$, 
$F_2=64$, 
and $F_3=64$. Image backbones used 4 convolutional building blocks with 32 channels. The threshold for calculating $\cc$ was set to $\epsilon=10^{-1}$. As for meta training, the gradient update step of inner-loop adaption was set to $10$, and we fixed $\alpha=10^{-2}$ and $\beta=10^{-4}$ in all experiments. For the BRNN in TFAN module, we chose to use a BiLSTM network to perform the iterative feature aggregation.  

We evaluated HetMAML and baselines based on their performance on the few-shot classification accuracy, and, to evaluate the efficiency of our model, we also report the meta-training time and the size of trainable meta-parameters.

\subsubsection{Task-Heterogeneous Few-Shot Classification}

Tables \ref{results_modelnet} and \ref{results_cub} report classification accuracy on each dataset. 
One can observe that HetMAML outperforms baselines on each dataset in terms of the classification accuracy within a few gradient updates. This demonstrates that HetMAML can successfully handle these heterogeneous task distribution cases and can fast adapt to all types of new tasks.  
The results on \textit{het}ModelNet-1 where all tasks are unimodal show that HetMAML can achieve the best performance in the cross-modal case of task heterogeneity. On the other datasets which have multimodal tasks, or especially with larger classes or less shots, HetMAML and Multi-MAML (BF) outperforms other baselines, showing that the proposed BRNN-based iterative feature aggregation (IFA) can achieve a proper balance between the efficiency of adaption and effectiveness of multimodal integration so that is suitable for few-shot problems. Multi-MAML (TF) which optimizes a large multimodal fusion network failed in few-shot learning scenarios as it requires a much larger parameter set due to its high-dimensional tensor. Instead, HetMAML encodes multi-modalities in a recurrent manner with a relatively small size of parameters. We also observe that the other five homogeneous baselines failed in all task-heterogeneity settings because they cannot customize the globally shared model architecture for tasks having different input structures. In contrast, HetMAML can better balance customization and generalization since the attentional contextual learning (ACL) can automatically take into account the context of tasks' input structures.

In addition, HetMAML outperforms Multi-MAML (BF), especially in 5/10-way 1-shot settings. 
It is because HetMAML simultaneously trains all types of tasks and thus can access more data from other types of tasks during training. In contrast, each MAML in Multi-MAMLs only uses the smaller number of tasks of a specific type for training these networks. Jointly training heterogeneous tasks could capture more reliable meta-knowledge rather than separately training multiple MAMLs.

\subsubsection{Adaption Speed Comparison} 
\begin{figure*}[t]
\centering
\includegraphics[width=2.07\columnwidth]{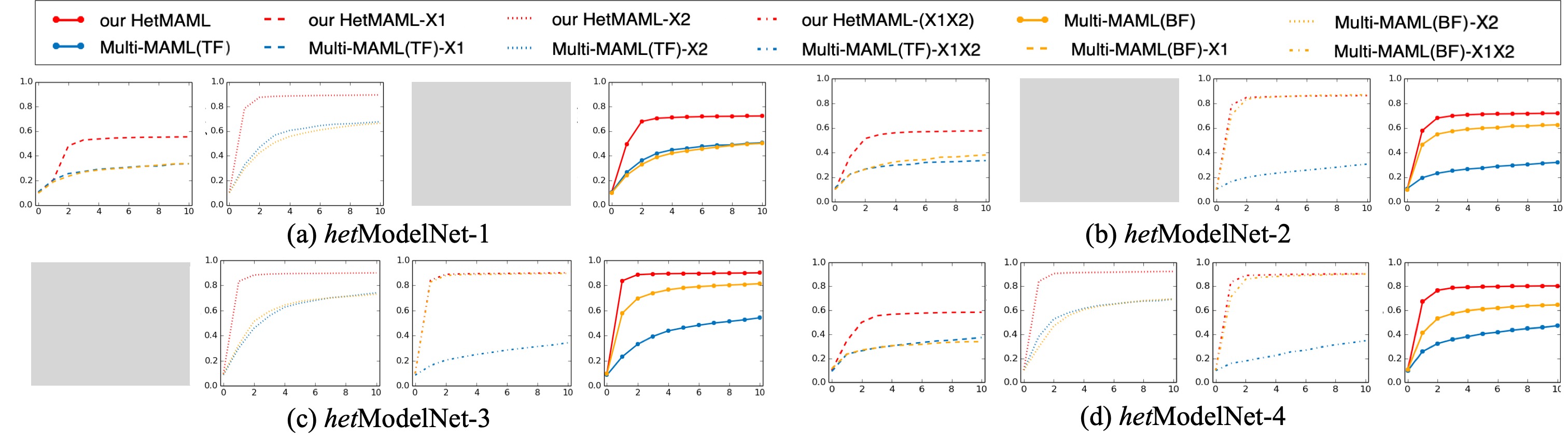}
\vspace{-0.15in}
\caption{Adaption speed comparison with 10-way 1-shot setting on \textit{het}ModelNet-1 (2, 3, and 4). For each dataset, the left three sub-figures indicate the results on each type of tasks, the right sub-figure indicates the average result of all testing tasks.
}
\label{appendix_exp}
\end{figure*}

We also compare adaption speed in Figure \ref{appendix_exp}, where we display the 10-way 1-shot classification accuracy (meta-test) with respect to the number of gradient updates, separately for each type of tasks on each of the four \textit{het}ModelNet datasets. 
In this figure, we compare HetMAML with Multi-MAML baselines on two datasets. Other baselines are not compared here because they view different types of tasks as the same type. 
Solid lines denote the average results of all types of tasks; dash lines denote the results of tasks that only have modality X1 as the input (type-X1); dotted lines denote the results of tasks that only have modality X2 as the input (type-X2); and dash-dotted lines denote tasks that have a pair of modalities as the input (type-(X1,X2)).


Overall, while the baselines need 8-9 or more gradient updates to adapt to new tasks, HetMAML can adapt to new tasks with just 2-3 updates, which demonstrates the fast adaption ability of HetMAML. 
We can observe that although HetMAML trains all types of tasks simultaneously, it can successfully fast adapt to each type of new tasks, suggesting that the proposed model is aware of the task-specific input structure. In addition, HetMAML outperforms baselines, especially on unimodal tasks, because the training of each types of tasks can benefit from more data from other types. Furthermore, these results are achieved with less training time and parameter size, showing that HetMAML is a more efficient framework than Multi-MAML baselines.
The adaption speed of the TFN-based baseline on type-(X1,X2) tasks is very slow (see the blue dash-dotted lines), suggesting that the large-scale multimodal fusion strategy is not suitable to few-shot multimodal tasks. According to the left-2 and left-3 charts, Type-(X1,X2) adaption in Multi-MAML (BF) is faster than the type-X1 or type-X2 adaption, proving the efficiency of iterative aggregation on few-shot multimodal fusion.

\begin{table}
\centering
  \begin{tabular}{p{3.5cm}cc}
  \toprule Method &  \text{\tabincell{c}{MT-Time}}  & \text{\tabincell{c}{Model Size}} \\ 
  \midrule
    Multi-MAML (TF) & 26.2 min & 2,762,905 \\
    Multi-MAML (BF) & 34.5 min & 1,736,783 \\
    \midrule
    \textbf{HetMAML} (ours) &\textbf{ 17.5 min} & \textbf{908,485} \\
\bottomrule
\end{tabular}
\caption{Comparison of model size and the average meta-training time (MT-Time) on \textit{het}ModelNet-4.}
\vspace{-0.1in}
\label{complexity}
\end{table}

\subsubsection{Efficiency and Scalability}
Table \ref{complexity} reports the comparison on the model sizes and training time between HetMAML and Multi-MAML. Compared with separate training baselines, our HetMAML uses one-third or a half of trainable meta-parameters to achieve better adaption performance, showing that HetMAML can efficiently and effectively generalize both type-specific and shared meta-parameters across heterogeneous tasks.
Moreover, HetMAML can be more scalable than baselines in terms of the number of data sources, in spite that our datasets include only two modalties because existing multimodal datasets having more modalities do not have enough samples to construct meta datasets. For HetMAML, the increase of $\triangle M$ modalities will result in  $\triangle M$ feature extractors $\{\theta_m|\forall m \in \triangle M\}$ added to the multi-channel backbone, while the other two modules' parameters will remain the same.  

\section{Related Works}

In this section, we briefly review related works in meta-learning, cross-modal domain adaption, multimodal multi-task learning, and multimodal integration.

\textbf{Meta-Learning}. Meta-learning approaches to few-shot learning problem mainly include gradient-based methods \cite{finn2017model,yoon2018bayesian,DBLP:conf/iclr/MishraR0A18,zintgraf2019fast} and metric-based methods \cite{snell2017prototypical,oreshkin2018tadam,koch2015siamese,vinyals2016matching,sung2018learning}. While \textit{metric-based} meta-learning learns 
a distance-based prediction rule over the embeddings, \textit{gradient-based} methods employ a bilevel learning framework that adapts the embedding model parameters given a task's training examples.
We follows gradient-based paradigm as the bilevel framework has the potential mechanisms for feature selection and could be more robust to input variations 
rather than metric-based frameworks.
Recent FSL \cite{xing2019adaptive,pahde2020multimodal,eloff2019multimodal,pahde2019self} 
extended to multimodal few-shot scenarios. Yet most of these methods assume homogeneous multimodal features, where input samples over tasks consist of the same set of modalities. 
In contrast, we focus on the case that some modalities may not be available in some tasks and thus different tasks may have different combinations of modalities. 
%
Recently, task-adaptive meta-learning \cite{suo2020tadanet,vuorio2019multimodal,kang2018transferable,DBLP:conf/iclr/RusuRSVPOH19} considered task distribution across domains, but different tasks still share the same input space. Different from these methods that assume task homogeneity, we study a novel FSL approach under heterogeneous task distributions.  

\textbf{Cross-modal Domain Adaption}. Domain Adaptation (DA) is useful in the situations where the source and target domains have the same classes but the input distribution of the target task is shifted with respect to the source task. Cross-modal DA deals with the special case that the inputs of the source and target differ in modalities \cite{pereira2014cross}, which is similar to the cross-modal case of our task heterogeneity. 
However, cross-modal DA focuses on the adaption between two domains (source and target). In contrast, this paper learns meta-knowledge for unlimited unseen tasks. There is no meta-objective in DA that optimizes ``how to learn" across tasks.

\textbf{Multimodal Multi-Task Learning}. Multi-Task Learning (MTL) aims to jointly learn several related tasks such that each task can benefit from parameters or representation sharing across different tasks. Multimodal MTL \cite{parthasarathy2017jointly,chen2017multimodal} 
deals with using MTL methods to solve multimodal integration problem.
Our HetMAML also attempts to design a parameter-sharing strategy so that all types of tasks can be learned with a unified framework. 
However, we aim to obtain meta-knowledge across unlimited tasks leveraging these shared parameters, whereas MTL intends to solve a fixed number of known tasks through a single-level optimization without a meta-objective.

\textbf{Multimodal Integration}. 
Deep multimodal integration studies how to align or integrate different modalities (e.g., visual, language, acoustic, or others) into a joint embedding. Existing methods include early fusion \cite{perez2013utterance,morency2011towards, snoek2005early,gunes2005affect}, late fusion that models the dynamics of multimodal interactions  ~\cite{zadeh2017tensor,liu2018efficient,chen2020hgmf}, and multimodal sequential learning \cite{zadeh2018memory,song2012multimodal,chambon2018deep}.
A majority of existing multimodal fusion models rely on large-scale training data so that their performance may drop dramatically in the few-data scenario. Also, most works build frameworks designed for single-type tasks, which are not very robust to heterogeneous few-shot tasks.
Although one can impute unavailable modalities using imputation strategies \cite{cai2018deep,tran2017missing,shang2017vigan,du2018semi} to convert the problem into homogeneity, this may introduce extra noise to the original few-data task, and also, the multimodal integration function may still vary between different types of tasks. In contrast, HetMAML directly learns with different input spaces and can be aware of the input structures of multimodal tasks to promote knowledge customization.

\section{Conclusions}

In this paper, we introduced HetMAML, a novel task-heterogeneous meta-agnostic meta-learning approach to few-shot learning under heterogeneous task distribution. To effectively capture  both  the  type-specific  and globally shared  meta-parameters, we designed a three-module framework, including a multi-channel backbone to extract modality-specific embeddings, a task-aware feature aggregation network to adaptively transform these multimodal embeddings into a task-specific representation by leveraging the task' type-specific knowledge, and a single-channel head module for decision making. Experiment results show that HetMAML is able to effectively and efficiently capture meta-parameters across heterogeneous tasks, balance customization and generalization, and successfully fast adapt to all types of new tasks.

\newpage
\bibliographystyle{ACM-Reference-Format}
\bibliography{sample-base}

\end{document}